\documentclass[sigconf]{acmart}

\AtBeginDocument{%
  \providecommand\BibTeX{{%
    \normalfont B\kern-0.5em{\scshape i\kern-0.25em b}\kern-0.8em\TeX}}}

\setcopyright{acmcopyright}
\copyrightyear{2022}
\acmYear{2022}
\acmDOI{XXXXXXX.XXXXXXX}

\acmConference[CIKM2022]{ACM International Conference on Information and Knowledge Management}{October 17-22, 2022}{Atlanta, Georgia, USA}
\begin{document}

\title{Time Series Anomaly Detection in Smart Homes:\\A Deep Learning Approach}

\author{Somayeh Zamani}
\affiliation{%
 \institution{University of Siegen}
 \streetaddress{Rono-Hills}
 \city{Siegen}
 \country{Germany}}
\email{somayeh.zamani@uni-siegen.de}

\author{Hamed Talebi}
\affiliation{%
  \institution{Amirkabir University of Technology}
  \streetaddress{30 Shuangqing Rd}
  \city{Tehran}
  \country{Iran}}
\email{hamed.talebi.aut@gmail.com}

\author{Gunnar Stevens}
\affiliation{%
  \institution{University of Siegen}
  \streetaddress{8600 Datapoint Drive}
  \city{Siegen}
  \country{Germany}
  \postcode{78229}}
\email{gunnar.stevens@uni-siegen.de}


\begin{abstract}
Fixing energy leakage caused by different anomalies can result in significant energy savings and extended appliance life. Further, it assists grid operators in scheduling their resources to meet the actual needs of end users, while helping end users reduce their energy costs. In this paper, we analyze the patterns pertaining to the power consumption of dishwashers used in two houses of the REFIT dataset. Then two autoencoder (AEs) with 1D-CNN and TCN as backbones are trained to differentiate the normal patterns from the abnormal ones. Our results indicate that TCN outperforms CNN1D in detecting anomalies in energy consumption. Finally, the data from the Fridge\_Freezer and the Freezer of house No. 3 in REFIT is also used to evaluate our approach.
\end{abstract}



\maketitle

\section{Introduction}
Throughout recent years, the energy demand has significantly gone up due to urban and industrial development alongside an increase in population \cite{pothitou2016framework}. Therefore, climate change, global warming, and volatility in energy prices have fuelled the interest in smart systems \cite{marikyan2019systematic}. In this regard, the huge potential increase of replacing traditional home appliances with new in-operation power-consuming ones by 2040 has caused the residential sector to account for roughly 60\% over 2017-25 and 70\% over 2025-40 of electricity demand increase of buildings. As such, household appliances need to be operating efficiently and used appropriately to achieve energy-saving goals \cite{hosseini2020practical}.


To this end, utilizing AI-based technologies and smart homes as novel interventions to recognize abnormal power utilization and understand the reasons for each abnormality could pave the way for end-consumers both to renovate wasteful devices and adopt a more sustainable energy consumption behavior \cite{schwartz2013cultivating}\cite{himeur2021artificial}\cite{chen2017butler}\cite{zhou2016smart}. \mbox{Moreover}, it facilitates the prediction of end-users power demand as well as performing an optimal energy distribution by grid operators depending on specific end-users’ needs. In addition, electrical anomalies are less likely to remain unnoticed for a long period of time which would result in higher power consumption or damage in the most critical cases \cite{castangia2021detection}.

Thus, for optimization purposes in smart homes, via implementing load monitoring systems and formulating smart anomaly detection models using machine learning techniques, the abnormality can be mitigated \cite{alsalemi2020appliance}. To do so, it is essential to analyze the energy consumption of households in order to identify consumption patterns and extract valuable information from smart homes \cite{castangia2021detection}\cite{das2009detecting}.

In this paper, the power consumption patterns of dishwashers used in houses No. 1 and 2 of the REFIT dataset are analyzed as examples of devices that are used based on the needs of users. Then, the data for each usage of the device is divided into different signals to properly train autoencoders with different backbones, including 1-dimensional CNN and TCN, to detect abnormal usage. For this purpose, any predicted value that is greater than twice the standard deviation of the electricity consumption the day before is considered abnormal.

The rest of the paper is organized as follows. We provide related work on anomaly detection in energy consumption in Section 2. In Section 3, our methodology is presented in details. Section 4 concludes the paper and discusses future work.
\begin{figure}[h]
  \centering
  \includegraphics[width=\linewidth]{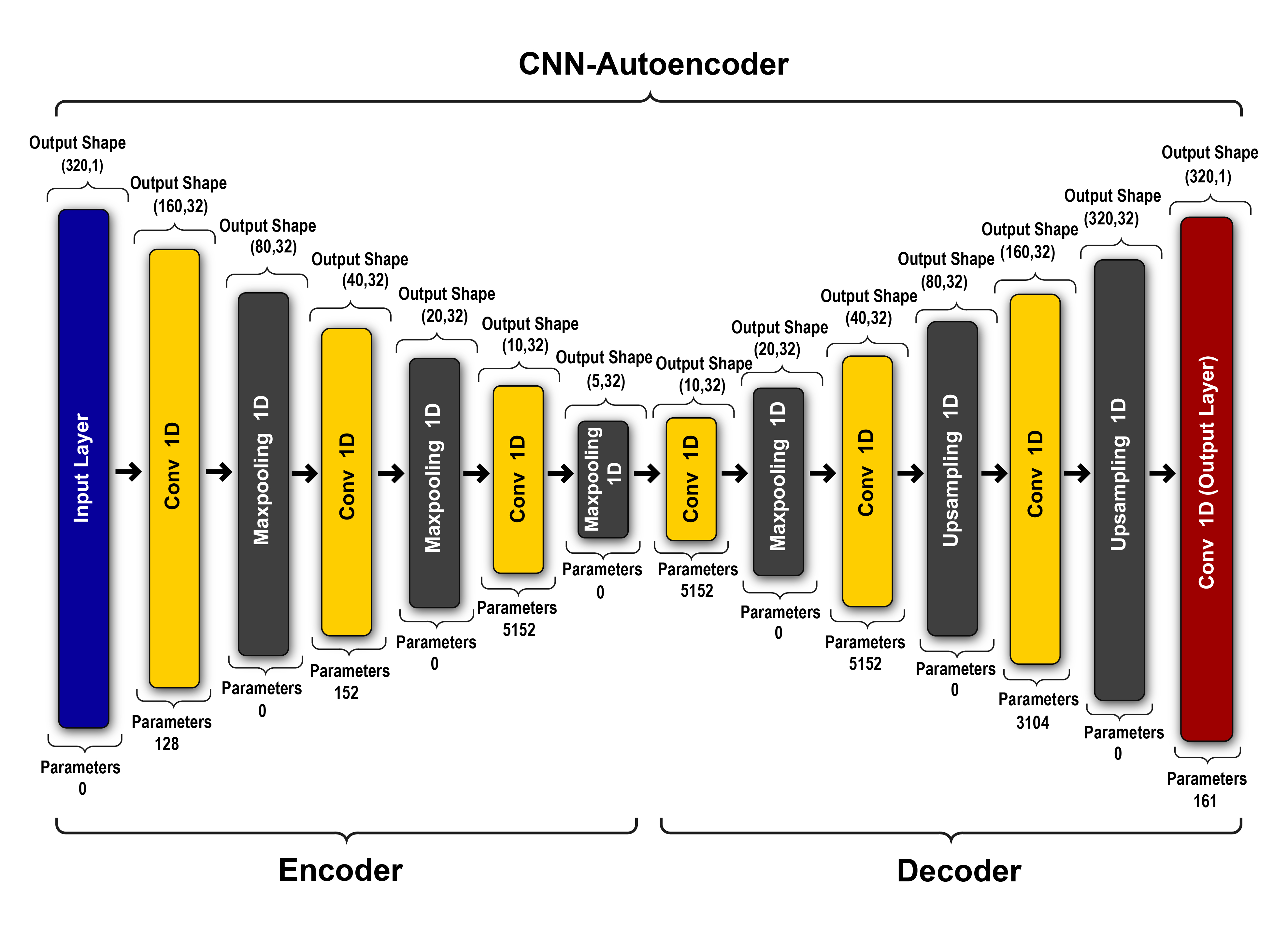}
  \caption{The architecture of the CNN-based autoencoder (CNN-AE)}
\end{figure}
\section{Related work}
In the context of energy usage, anomalies are defined as deviations from expected behavior that occur when the consumption of a household appliance does not correspond with its normal pattern \cite{hosseini2020practical}\cite{pang2021deep}. Among the key applications of anomaly detection by load monitoring, are forecasting maintenance and energy efficiency \cite{ullah2017predictive}. Thus, a smart plug, smart appliance, and other appliance-level monitoring devices are needed to continuously monitor the power consumption of individual appliances in a house \cite{castangia2021detection}. However, identifying anomalies, and their nature of them should also be considered, which can be categorized, based on different dimensions. In the data science world, anomalies are seen as either single points that are not necessarily relevant to each other or a set of data points that constitute a pattern and, therefore, can be interpreted in relation to each other.
The other dimension of anomaly detection that should be taken into account is the context which refers to a deviation in a particular context relating to the structure of the data \cite{hosseini2020practical}\cite{chandola2009anomaly}\cite{song2007conditional}. For example, in the context of a warm season, a temperature report of -30 degrees Celsius can be anomalous; however, during a cold season, such a report may be more common \cite{hosseini2020practical}.
To this end, understanding the available data will provide a solid foundation for improving energy efficiency. For this purpose, there are thirty-one publicly available databases with several features, such as the geographical location, period of collection, number of monitored households, the sampling rate of collected data, and number of sub-metered appliances \cite{himeur2020building}. Regarding this, a valuable dataset is REFIT which includes cleaned electrical consumption in Watts for 20 households in the UK at both the aggregate and appliance levels \cite{murray2017electrical}.
On the other hand, Pang \cite{pang2021deep} has provided a comprehensive overview of current anomaly detection methods to gain an important understanding of their inherent capabilities and limitations in addressing some largely unsolved challenges in anomaly detection.
According to his study, Autoencoders, which are a subset of the generic normality feature learning category, aim to learn some low-dimensional feature representation space on which the given data instances can be well reconstructed. While this is a widely used method for data compression or dimension reduction, by using this method, the feature representations are enforced to learn important regularities of the data so that reconstruction errors are minimized. Consequently, anomalies are difficult to reconstruct from the resulting representations and are, therefore, subject to large reconstruction errors \cite{hasan2016learning}.

\section{Methodology}
\subsection{Dataset and preprocessing}
The REFIT Electrical Load Measurements dataset contain cleaned electrical consumption data in Watts for 20 households in the UK at both the aggregate and appliance level. The data is related to a period of two years comprising nine individual appliance measurements at 8-second intervals per house with 1,194,958,790 readings \cite{murray2017electrical}. The models proposed in this paper are trained using dishwasher data from houses No. 1 and 2. Furthermore, data from the Fridge\_Freezer and the Freezer of house No. 3 is used to assess the effectiveness of our approach.

To begin with, it is necessary to resample the data to convert it into equal time intervals {\itshape r}. Then using the following formula, the average sampling time, $\overline{t}$ of the REFIT data is used to fill in a limited number of signals, {\itshape n} with no data. The remaining empty intervals are substituted with zero.
\begin{equation}
    n = \left[ \frac{4*\overline{t}}{r} \right]
\end{equation}
Additionally, for devices used according to users' needs, consumption data must first be differentiated. The power consumption pattern may include turning the device on and off several times per usage. The matching data is therefore combined into relevant signals. Also, due to the possibility of failure in some devices that can result in constant operation for an extended period, we assume a maximum period for a device to operate.
\subsection{The proposed models}
The development of time series anomaly detection algorithms has recently received considerable attention. Autoencoder-based approaches are often used to identify anomalous behavior by analyzing the reconstruction error of the data \cite{Hawkins2002}\cite{Malhotra2016}\cite{Thill2020}. Having learned a nonlinear transformation of the input data into a compressed representation, latent variables are used to reconstruct the original input.
On the other hand, utilizing the convolution mechanism in sequential models is computationally optimal \cite{kim2014convolutional}. Also, due to CNNs' equivariance properties and sparse interactions, they are translated from computer vision into the time domain using temporal convolutional networks (TCN)\cite{Thill2020}.

In the following sections, we will describe how we used autoencoders (AEs) for time series data that utilize 1-dimensional CNNs and TCNs as building blocks to detect energy anomalies in the REFIT dataset.
\begin{figure}[h]
  \centering
  \includegraphics[width=\linewidth]{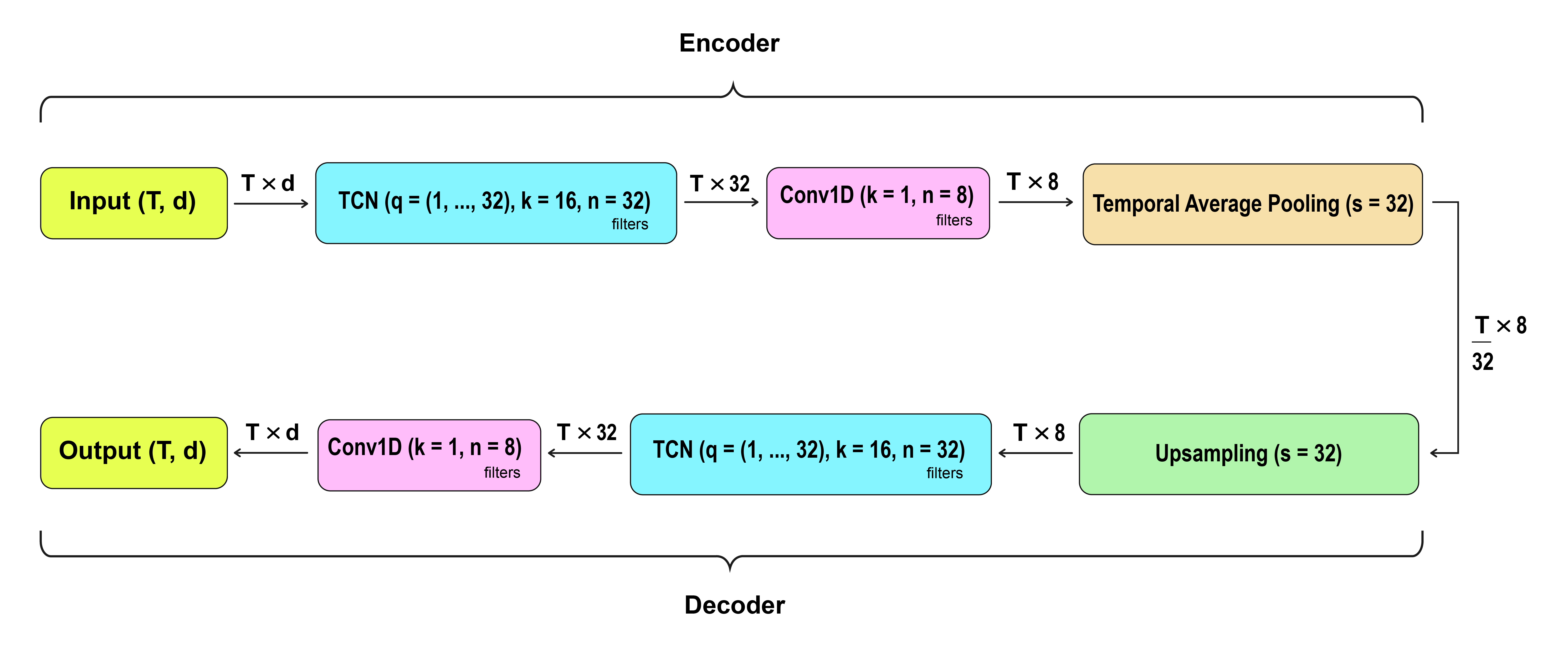}
  \caption{The architecture of the TCN-based autoencoder (TCN-AE)}
\end{figure}
\subsubsection{CNN-based autoencoder (CNN-AE)}
We used TensorFlow to implement the architecture consisting of two smaller sequential models, an encoder and a decoder. Also, considering the speed of the model convergence, our CNN-based autoencoder is comprised of 3 layers of Conv1D using the data of the households’ dishwashers. Furthermore, a nonlinear ReLu activation function is used in each convolution layer. In this model, a standard rate of 0.2 is considered for the dropout layer to randomly remove 20\% of the upper layer during learning.

Figure 1. shows the layers and the number of input and output parameters of each.
The input layer is $320\times1$ (3200 seconds), calculated according to the maximum operation time of the device.
\subsubsection{TCN-based autoencoder (TCN-AE)}
The temporal convolutional network (TCN) combines simplicity with auto-regressive prediction, residual blocks, and a very long memory. In general, a TCN can be broken down into three components: a list of dilation rates $D=\{{q_1, q_2, ..., q_{n_{r}}}\}$, the number of $n_{filters}$, and the kernel size $k$, which is the same for all filters in a TCN \cite{Thill2020}\cite{Bai2018}.
Inspired by a classical (deep) autoencoder, the TCN autoencoder encodes sequences, along the temporal axis, of length $T$ into a compressed representation of length $T/s$ (where ${s\in\mathbb{Z}^+}$) and then tries to reconstruct the original sequence\cite{Thill2020}.




In Figure 2, each layer of the TCN-AE is described by its parameters within the box. TCN-AE receives a sequence $x[n]$ of length $T$ and dimensionality $d$ as its input. Using a TCN, the encoder first processes input sequence $x[n]$ of length $T$ and dimension $d$. Afterward, a one-dimensional convolutional layer with $q= 1$, $k= 1$, and $n_{filters} = 8$ is used to reduce the dimensionality of the TCN's output. As the last layer in the encoder, the temporal average pooling layer downsamples the series by a factor of $s$. To do so, groups of size $s$ are averaged along the time axis.

In the decoder module, the downsampled sequence is returned to its original length by performing a nearest neighbor interpolation on the upsampled sequence. Upsampled sequences are passed through a second TCN with independent weights parameterized similarly to the encoder-TCN. As a final step, the input sequence is reconstructed with a Conv1D layer that ensures that the dimensionality of the input is matched (by setting $k = 1$ and $n_{filters} = d$) \cite{Thill2020}. As described in the next section, the input sequence and its reconstruction will be used for detecting anomalies after TCN-AE has been trained.

\begin{figure}[h]
  \centering
  \includegraphics[width=\linewidth]{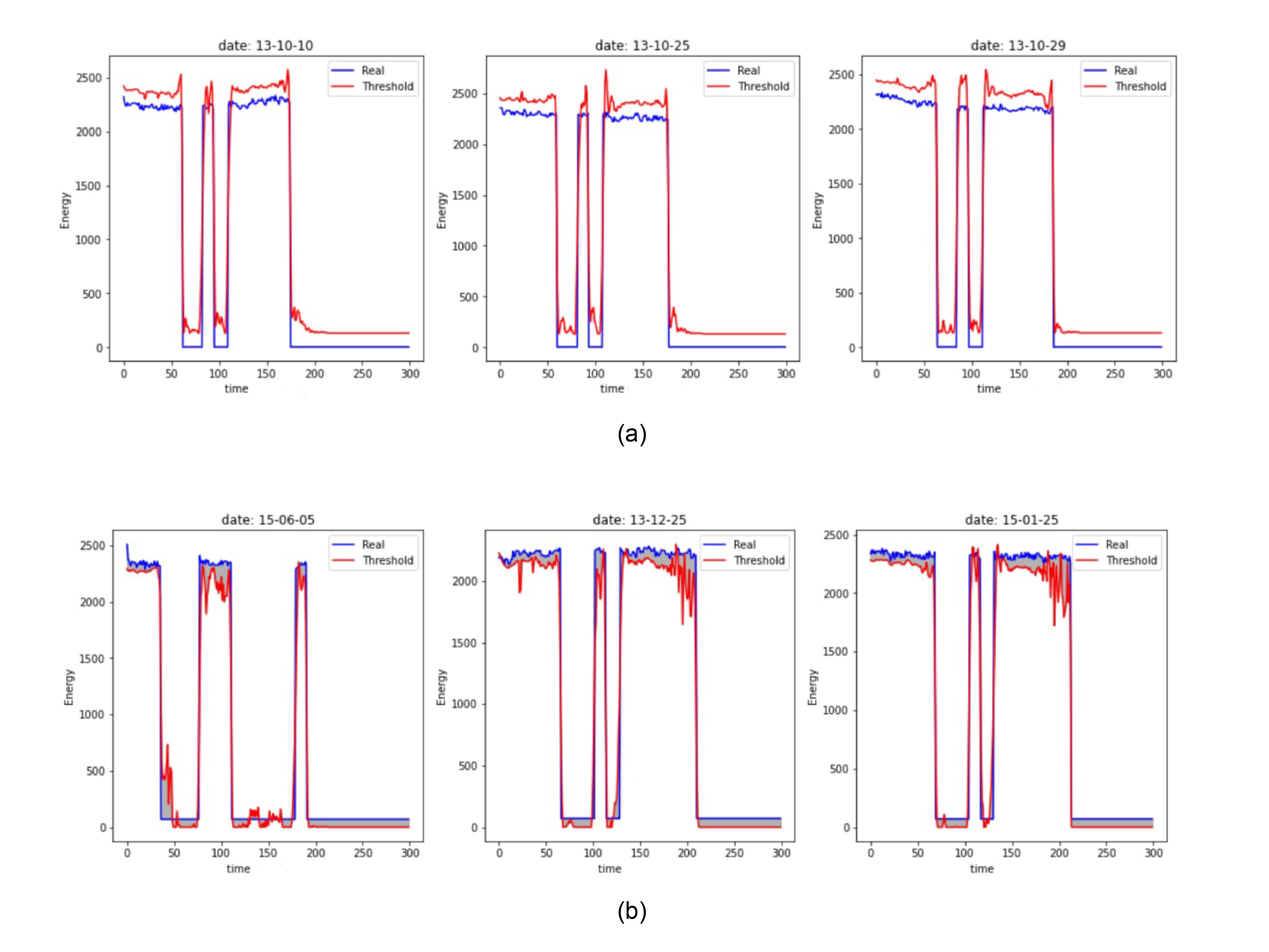}
  \caption{(a) Examples of the normal energy consumption of the dishwasher (b) Examples of the abnormal energy consumption of the dishwasher}
\end{figure}
\subsection{Experimental results}
\subsubsection{Anomaly detection}
We compute a threshold value of $2\sigma$ above the predicted value to measure the trend in electricity consumption over time, where $2\sigma$ is the standard deviation on the day before the actual moment \cite{Pan2022}. An abnormal state is defined as a value exceeding the threshold for predicted electricity consumption at the actual moment. Equations (2) and (3) show the calculation of $\sigma$ and $\gamma_{threshold}$:
\begin{equation}
    \sigma=\sqrt{\frac{\sum_{i=1}^{n} (x_i - \overline{x})^2}{n}}
\end{equation}
\begin{equation}
    \gamma_{threshold} = \widehat{\gamma} + 2\sigma
\end{equation}
where $\sigma$ is the standard deviation, $\gamma_{threshold}$ is the threshold, $\widehat{\gamma}$ is the predicted value, $x_i$ is the electricity consumption, $\overline{x}$ is the average electricity consumption, and $n$ is the number of samples.

A normal electricity usage pattern detection for the dishwasher is shown in Figure 3 (a), where the real-time threshold curve follows the sequence trend, indicating the model depicts the dishwasher's normal electricity usage. As can be seen in the figure, the real power consumption curve does not exceed the threshold range, which indicates a normal level of electricity consumption. As shown in Figure 3 (b), anomalous consumption patterns occur when actual values exceed the threshold. Consequently, the method can distinguish between normal and abnormal consumption behavior.


\begin{table*}
  \caption{Evaluation of CNN1D-AE performance with a different data division ratio}
  \label{tab:commands}
  \begin{tabular}{ccccc}
    \toprule
    Model &Division Ratio & MAE & MAPE\\
    \midrule
    \texttt{CNN1D-AE} & 9:1& 0.1781 & \%22.55\\
    \texttt{CNN1D-AE} & 8:2 & \textbf{0.1570} & \textbf{\%20.07}\\
    \texttt{CNN1D-AE} & 7:3& 0.1702 & \%21.99\\
    \bottomrule
  \end{tabular}
\end{table*}

\begin{table*}
  \caption{Evaluation of TCN-AE performance with a different data division ratio}
  \label{tab:commands}
  \begin{tabular}{ccccc}
    \toprule
    Model &Division Ratio & MAE & MAPE\\
    \midrule
    \texttt{TCN-AE} & 9:1& 0.1527 & \%21.39\\
    \texttt{TCN-AE} & 8:2 & \textbf{0.1371} & \textbf{\%17.52}\\
    \texttt{TCN-AE} & 7:3& 0.1412 & \%19.88\\
    \bottomrule
  \end{tabular}
\end{table*}

\begin{table*}
  \caption{Evaluation of the model's performance using the Fridge\_Freezer and the Freezer data of house \#3}
  \label{tab:commands}
  \begin{tabular}{ccccc}
    \toprule
    Model &Division Ratio & MAE & MAPE\\
    \midrule
    \texttt{CNN1D-AE} & 9:1& 0.1678 & \%21.15\\
    \texttt{CNN1D-AE} & 8:2 & \textbf{0.1486} & \textbf{\%19.17}\\
    \texttt{CNN1D-AE} & 7:3& 0.1649 & \%20.29\\
    \hline \hline 
    \texttt{TCN-AE} & 9:1& 0.1422 & \%18.21\\
    \texttt{TCN-AE} & 8:2 & \textbf{0.1264} & \textbf{\%16.33}\\
    \texttt{TCN-AE} & 7:3& 0.1353 & \%17.70\\
    \bottomrule
  \end{tabular}
\end{table*}
\subsubsection{Evaluation}
As Table 1. and Table 2. show, for both architectures, the best performance is obtained with the data division ratio of 8:2, and clearly, TCN-AE is more efficient than CNN-AE. Our unsupervised approach has also been evaluated using the data from the Fridge\_Freezer and the Freezer of house No. 3 in REFIT. The results in Table 3. confirm the best division ratio of 8:2 and the higher performance of TCN compared to CNN1D.

\section{Conclusion and future work}
This paper presents the starting point of our work on studying how would applying deep learning algorithms, and explainability improve energy efficiency, environmental sustainability, and user adoption. In this regard, first, we preprocessed our data by resampling and differentiating each usage. Next, the extracted patterns of dishwasher usage in houses No. 1 and 2 of the REFIT dataset were analyzed. Two deep learning models, CNN-AE and TCN-AE were then trained to detect abnormalities. While the TCN backbone performed better, we evaluated our models using the data from the refrigerators of house No. 3 in REFIT as well.

Through the implementation of energy monitoring systems and the formulation of intelligent anomaly detection techniques, abnormal behaviors can be mitigated. This is possible, especially if user-centric explainable recommender systems are combined with anomaly detection modules. However, there is no proper labeled dataset available to develop accurate algorithms or detect different types of anomalies. Accordingly, we plan to run a laboratory to build the first appropriately labeled energy anomaly dataset.
\section{Acknowledgement}
This research has been funded by the EU Horizon 2020 Marie Skłodowska-Curie International Training Network GECKO, Grant number 955422.



\bibliographystyle{unsrt}
\bibliography{main}

\begin{thebibliography}{10}

\bibitem{pothitou2016framework}
Mary Pothitou, Athanasios~J Kolios, Liz Varga, and Sai Gu.
\newblock {A framework for targeting household energy savings through habitual
  behavioural change}.
\newblock {\em International Journal of Sustainable Energy}, 35(7):686--700,
  2016.

\bibitem{marikyan2019systematic}
Davit Marikyan, Savvas Papagiannidis, and Eleftherios Alamanos.
\newblock {A systematic review of the smart home literature: A user
  perspective}.
\newblock {\em Technological Forecasting and Social Change}, 138:139--154,
  2019.

\bibitem{hosseini2020practical}
Sayed~Saeed Hosseini, Kodjo Agbossou, Sousso Kelouwani, Alben Cardenas, and
  Nilson Henao.
\newblock {A practical approach to residential appliances on-line anomaly
  detection: A case study of standard and smart refrigerators}.
\newblock {\em IEEE Access}, 8:57905--57922, 2020.

\bibitem{schwartz2013cultivating}
Tobias Schwartz, Sebastian Denef, Gunnar Stevens, Leonardo Ramirez, and Volker
  Wulf.
\newblock {Cultivating energy literacy: results from a longitudinal living lab
  study of a home energy management system}.
\newblock In {\em Proceedings of the SIGCHI Conference on Human Factors in
  Computing Systems}, pages 1193--1202, 2013.

\bibitem{himeur2021artificial}
Yassine Himeur, Khalida Ghanem, Abdullah Alsalemi, Faycal Bensaali, and Abbes
  Amira.
\newblock {Artificial intelligence based anomaly detection of energy
  consumption in buildings: A review, current trends and new perspectives}.
\newblock {\em Applied Energy}, 287:116601, 2021.

\bibitem{chen2017butler}
Siyun Chen, Ting Liu, Feng Gao, Jianting Ji, Zhanbo Xu, Buyue Qian, Hongyu Wu,
  and Xiaohong Guan.
\newblock {Butler, not servant: A human-centric smart home energy management
  system}.
\newblock {\em IEEE Communications Magazine}, 55(2):27--33, 2017.

\bibitem{zhou2016smart}
Bin Zhou, Wentao Li, Ka~Wing Chan, Yijia Cao, Yonghong Kuang, Xi~Liu, and Xiong
  Wang.
\newblock {Smart home energy management systems: Concept, configurations, and
  scheduling strategies}.
\newblock {\em Renewable and Sustainable Energy Reviews}, 61:30--40, 2016.

\bibitem{castangia2021detection}
Marco Castangia, Riccardo Sappa, Awet~Abraha Girmay, Christian Camarda, Enrico
  Macii, and Edoardo Patti.
\newblock {Detection of Anomalies in Household Appliances from Disaggregated
  Load Consumption}.
\newblock In {\em 2021 International Conference on Smart Energy Systems and
  Technologies (SEST)}, pages 1--6. IEEE, 2021.

\bibitem{alsalemi2020appliance}
Abdullah Alsalemi, Yassine Himeur, Faycal Bensaali, and Abbes Amira.
\newblock {Appliance-level monitoring with micro-moment smart plugs}.
\newblock {\em arXiv preprint arXiv:2012.05787}, 2020.

\bibitem{das2009detecting}
Kaustav Das.
\newblock {\em {Detecting patterns of anomalies}}.
\newblock Carnegie Mellon University, 2009.

\bibitem{pang2021deep}
Guansong Pang, Chunhua Shen, Longbing Cao, and Anton Van~Den Hengel.
\newblock {Deep learning for anomaly detection: A review}.
\newblock {\em ACM Computing Surveys (CSUR)}, 54(2):1--38, 2021.

\bibitem{ullah2017predictive}
Irfan Ullah, Fan Yang, Rehanullah Khan, Ling Liu, Haisheng Yang, Bing Gao, and
  Kai Sun.
\newblock {Predictive maintenance of power substation equipment by infrared
  thermography using a machine-learning approach}.
\newblock {\em Energies}, 10(12):1987, 2017.

\bibitem{chandola2009anomaly}
V~Chandola, A~Banerjee, and V~Kumar.
\newblock {Anomaly Detection: A Survey. ACM Computing Surveys}.
\newblock {\em vol}, 41:15, 2009.

\bibitem{song2007conditional}
Xiuyao Song, Mingxi Wu, Christopher Jermaine, and Sanjay Ranka.
\newblock {Conditional anomaly detection}.
\newblock {\em IEEE Transactions on knowledge and Data Engineering},
  19(5):631--645, 2007.

\bibitem{himeur2020building}
Yassine Himeur, Abdullah Alsalemi, Faycal Bensaali, and Abbes Amira.
\newblock {Building power consumption datasets: Survey, taxonomy and future
  directions}.
\newblock {\em Energy and Buildings}, page 110404, 2020.

\bibitem{murray2017electrical}
David Murray, Lina Stankovic, and Vladimir Stankovic.
\newblock {An electrical load measurements dataset of United Kingdom households
  from a two-year longitudinal study}.
\newblock {\em Scientific data}, 4(1):1--12, 2017.

\bibitem{hasan2016learning}
Mahmudul Hasan, Jonghyun Choi, Jan Neumann, Amit~K Roy-Chowdhury, and Larry~S
  Davis.
\newblock {Learning temporal regularity in video sequences}.
\newblock In {\em Proceedings of the IEEE conference on computer vision and
  pattern recognition}, pages 733--742, 2016.

\bibitem{Hawkins2002}
Simon Hawkins, Hongxing He, Graham Williams, and Rohan Baxter.
\newblock {Outlier detection using replicator neural networks}.
\newblock In {\em International Conference on Data Warehousing and Knowledge
  Discovery}. Springer, 2002.

\bibitem{Malhotra2016}
Pankaj Malhotra, Anusha Ramakrishnan, Gaurangi Anand, Lovekesh Vig, Puneet
  Agarwal, and Gautam Shroff.
\newblock {LSTM-based encoder-decoder for multi-sensor anomaly detection}.
\newblock {\em arXiv preprint arXiv:1607.00148}, 2016.

\bibitem{Thill2020}
Markus Thill, Wolfgang Konen, and Thomas B{\"{a}}ck.
\newblock {Time series encodings with temporal convolutional networks}.
\newblock In {\em International Conference on Bioinspired Methods and Their
  Applications}, pages 161--173. Springer, 2020.

\bibitem{kim2014convolutional}
Yoon Kim.
\newblock {Convolutional neural networks for sentence classification}.
\newblock {\em arXiv preprint arXiv:1408.5882}, 2014.

\bibitem{Bai2018}
Shaojie Bai, J~Zico Kolter, and Vladlen Koltun.
\newblock {An empirical evaluation of generic convolutional and recurrent
  networks for sequence modeling}.
\newblock {\em arXiv preprint arXiv:1803.01271}, 2018.

\bibitem{Pan2022}
Haipeng Pan, Zhongqian Yin, and Xianzhi Jiang.
\newblock {High-Dimensional Energy Consumption Anomaly Detection: A Deep
  Learning-Based Method for Detecting Anomalies}.
\newblock {\em Energies}, 15(17):6139, 2022.

\end{thebibliography}

\end{document}